\documentclass[a4paper,twoside]{article}

\usepackage{mystyle}

\begin{document}

\title{Single Stage Class Agnostic Common Object Detection: \\
	A Simple Baseline}

\author{\authorname{Chuong H. Nguyen\sup{1}, Thuy C. Nguyen \sup{1}, Anh H. Vo \sup{1}  and Yamazaki Masayuki \sup{2}}
\affiliation{\sup{1}Cybercore, Marios 10F, Morioka Eki Nishi Dori 2-9-1, Morioka, Iwate, Japan}
\affiliation{\sup{2}Toyota Research Institute-Advanced Development, 3-2-1 Nihonbashi-Muromachi, Chuo-ku, Tokyo, Japan}
\email{\{chuong.nguyen, thuy.nguyen, anh.vo\}@cybercore.co.jp, masayuki.yamazaki@tri-ad.global}
}


\abstract{
	This paper addresses the problem of common object detection, which aims to detect objects of similar categories from  a set of images. Although it shares some similarities with the standard object detection and co-segmentation, common object detection, recently promoted by \cite{Jiang2019a}, has some unique advantages and challenges. First, it is designed to work on both closed-set and open-set conditions, a.k.a. known and unknown objects. Second, it must be able to match objects of the same category but not restricted to the same instance, texture, or posture. Third, it can distinguish multiple objects. In this work, we introduce the Single Stage Common Object Detection (SSCOD) to detect class-agnostic common objects from an image set. The proposed method is built upon the standard single-stage object detector. Furthermore, an embedded branch is introduced to generate the object's representation feature, and their similarity is measured by cosine distance. Experiments are conducted on PASCAL VOC 2007 and COCO 2014 datasets. While being simple and flexible, our proposed SSCOD built upon ATSSNet performs significantly better than the baseline of the standard object detection, while still be able to match objects of unknown categories. Our source code can be found at \href{https://github.com/cybercore-co-ltd/Single-Stage-Common-Object-Detection}{(URL)}.
}

\onecolumn \maketitle \normalsize \setcounter{footnote}{0} \vfill

\section{\uppercase{Introduction}}
\noindent The ability to find similar objects across different scenes is important for many applications, such as object discovery or image retrieval. Different from the standard object detection, which can only make correct predictions on a close-set of predefined categories, common-object detection (COD) aims to locate general objects appearing in both scenes, regardless of their categories.     

The COD problem \cite{Jiang2019a} is closely related to co-segmentation, co-detection, and co-localization tasks. Although they all attempt to propose the areas belonging to common object categories, there are several key differences. Concretely, co-segmentation does not distinguish different instances. Co-detection finds the same object instance in a set of images, while co-localization is restricted to finding one category that appears in multiple images. The COD problem hence is much more challenging, due to (1) it must be able to localize potential areas containing objects, (2) be able to work in open-set condition, i.e. detect unseen categories, (3) be able to match those of same categories and not limited to the same instance, texture or posture.

\begin{figure}
	\centering
	\includegraphics[width=0.47\textwidth]{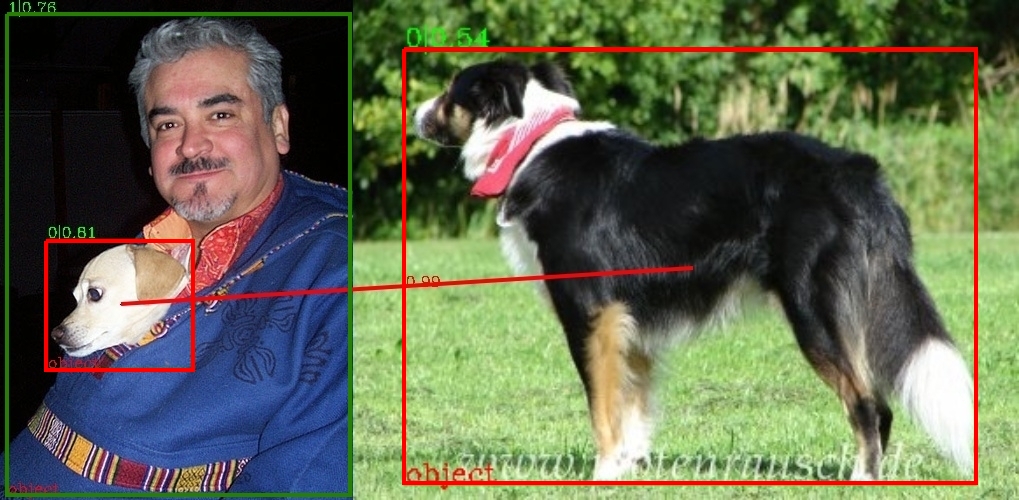} \hspace{0.05in}
	\includegraphics[width=0.47\textwidth]{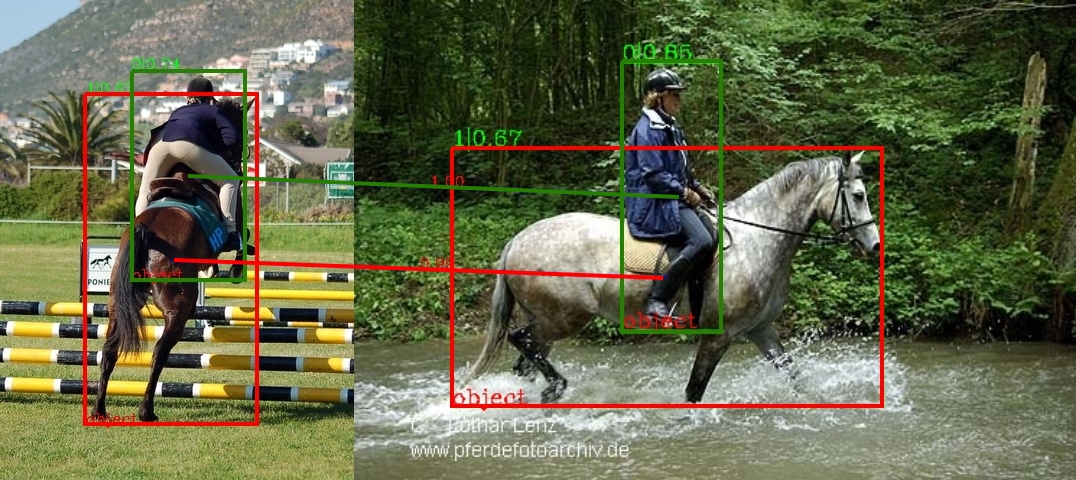}
	\caption{Illustrated results of common object detection.}
	\label{fig:Example}
\end{figure}

In this work, we focus on a similar COD problem, which is applied for 2D images domain, as illustrated in \fig{Example}. The objective is to find a set of bounding box pairs from two input images, such that each pair contains objects of the same category. Also, there should be no restriction on the number of classes, seen or unseen categories, and the number of instances in the images. Our direct application is to detect suspicious objects in surveillance cameras, hence the ability to detect unknown objects is critical. Moreover, since we need to perform detection in real-time, we select an
FPGA as our target hardware. This limits the type of kernels we can perform to
standard operations, i.e. some operators such as ROI-Align are not supported. Concretely, we introduce a Single Stage Common Object Detection (SSCOD), in which our contributions are summarized as follows:                                   

\begin{itemize}
	\item Single Stage Common Object Detector is proposed. The framework is simple and can be adapted to any standard Single Stage Object Detection. Hence, the standard training pipeline can be used, except that the classification branch is trained in a class-agnostic manner. This helps the network generalize the concept of objectness seen in training set to detect similar categories that are unseen.
	\item An embedded branch is introduced to extract an object's representation feature, using its cosine similarity to detect common object pairs. We investigate different loss functions for metric learning, such as classwise and pairwise losses, and then propose a unified function named Curriculum Contrastive Loss to deal with inherent problems of object detection, such as class imbalance and small batch-size. Our proposed loss yields the best results in our experiments.
	\item The model is evaluated on two dataset PASCAL VOC 2007 and COCO 2014. Our SSCOD model can achieve better results than the baseline of standard object detection, for both known and unknown categories. For unknown cases, our SSCOD can achieve comparable results with previous work. 
\end{itemize}  

\section{\uppercase{Literature Review}}
\subsection{Object Detection}
Since the COD problem is developed based on Object Detection framework, we review general techniques for Object detection in this section. Object Detection framework can be separated into two main approaches, namely two stages and one stage. 

Benchmarks in two-stage approach can be named as Regional-based convolutional neural network \cite{Girshick2014}, \cite{Girshick2015}, Faster R-CNN \cite{Ren2015} and Mask R-CNN \cite{He2017a}. This approach is based on a backbone CNN to extract features, which is then attached to two CNN modules. The first one proposes possible regions containing objects, and the second module contains two sub-nets: a classification head to classify the object and a regression head to predict bounding boxes offset from the anchor. Since Region Proposal is the core component, recent works, such as Iterative RPN \cite{Gidaris2016} or Cascade RCNN \cite{Cai2018b}, \cite{Cai2019}  attempt to enhance its performance by adding more stages to refine the predictions. Recently, Cascade RPN \cite{Vu2019} improves the quality of region proposal by using Adaptive Convolution and combine anchor-based and anchor-free criteria to define positive boxes. \cite{Song2020} improves the spatial misalignment between classification and regression heads by using two disentangled proposals, which are estimated by the shared proposal. In general, two-stage approaches can achieve higher accuracy by cascading more stages and refine modulators. However, it is often slower due to the framework complexity.

Single-stage detectors were developed later to improve speed, and the representative works can be named as SSD \cite{Liu2016}, YOLO \cite{Redmon2017}, \cite{Redmon2018}, and RetinaNet \cite{Lin2017}. The key advantage of this approach is to omit the proposal region but use a sliding-window to produce dense predictions directly. Specifically, at each cell in a feature map, a set of default anchors with different scales and ratios are predefined. Classification and bounding box regression are then predicted directly on each anchor. Recently, research shifts attention to remove the anchor-box step and propose a new kind of framework name ``anchor-box-free" approach. In general, anchor-box-free approaches, such as FCOS \cite{Tian2019}), CenterNet \cite{Duan2019a}, Object-as-Points \cite{Zhou2019}, CornerNet \cite{Law2018} are designed to be simpler and more efficient. \cite{Zhang2019c} investigates the factors constituting the performance gap between anchor and free-anchor approaches, and discover that the main factor is how to assign positive/negative training samples. Consequently, they propose adaptive training sample selection (ATSS) improvement to Retina net, which surpasses all anchor and free-anchor approaches without introducing any overhead. In short, single-stage detectors currently achieve comparable or even better accuracy than two-stages, while significantly simpler and faster. 

\subsection{Co-Segmentation}
The co-segmentation has been studied for many years \cite{Joulin2010}, \cite{Vicente2011} \cite{7780450}, where the main goal is to segment common foreground in the pixel level from multiple images. \cite{Yuan2017} introduced a deep dense conditional random field framework and used handcrafted SIFT and HOG features to establish co-occurrence maps. \cite{7780450} proposed a manifold ranking method that combines low-level appearance features and high-level semantic features extracted from an Imagenet pre-trained network.  

However, the application of previous works is quite restricted, since it assumes only a single common object, which also must be salient in the image set. Recently, \cite{Li2019d} proposed a deep Siamese network to achieve object co-segmentation from a pair of images. \cite{Chen2019b} proposed an attention mechanism to select areas that have high activation in feature maps for all input images. \cite{Zhang2019e} proposed a spatial-semantic modulated network, in which the spatial module roughly locates the common foreground by capturing the correlations of feature maps across images, and the semantic module refines the segmentation masks. A comprehensive review of co-segmentation methods can be found in the recent work of \cite{Xu2019}, \cite{Merdassi2019}. The co-segmentation setting, however, works with pixel-level rather than object instance level. Hence, the objective is different from the COD problem.
 
\subsection{Common Object Detection}
\cite{Bao2012} introduce a problem named object co-detection aiming to detect if the same object is present in a set of images. It is based on the intuition that an object should have a consistent appearance regardless of observation viewpoints. \cite{Guo2013a} follow the principle to exploit the consistent visual patterns from the objects. The goal then is to recognize whether objects in different images correspond to the same instance, and estimate the viewpoint transformation.         

Co-localization \cite{Le2017}, \cite{Li2019} defines the problem as localizing categorical objects using only a positive set of sample images. The general approach is to utilize a classification activation map from a pre-trained Imagenet network to localize the common areas. This problem is weaker since it requires a set of positive images as input, hence the application is limited to a single instance only.

\cite{Jiang2019a} recently extend the idea of co-detection and co-localization and introduce common object detection, which removes the aforementioned limitations. In their approach, Faster-RCNN is used as the base detector to propose foreground areas. The object proposals are then passed to an ROI align layer to extract object features in the second stage. Siamese Network and Relation Matching subnet are proposed to estimate the similarity between objects. Compared to \cite{Jiang2019a} solution, our proposed method is based on a Single-Stage Detector with an embedded branch network, which is more simple and flexible. Moreover, our proposed method can achieve higher accuracy in both seen and unseen categories, as presented in the following sections.

\section{\uppercase{Methodology}}
\subsection{Network Architecture}
\begin{figure*}[h]
 	\centering
 	\includegraphics[width=0.9\textwidth]{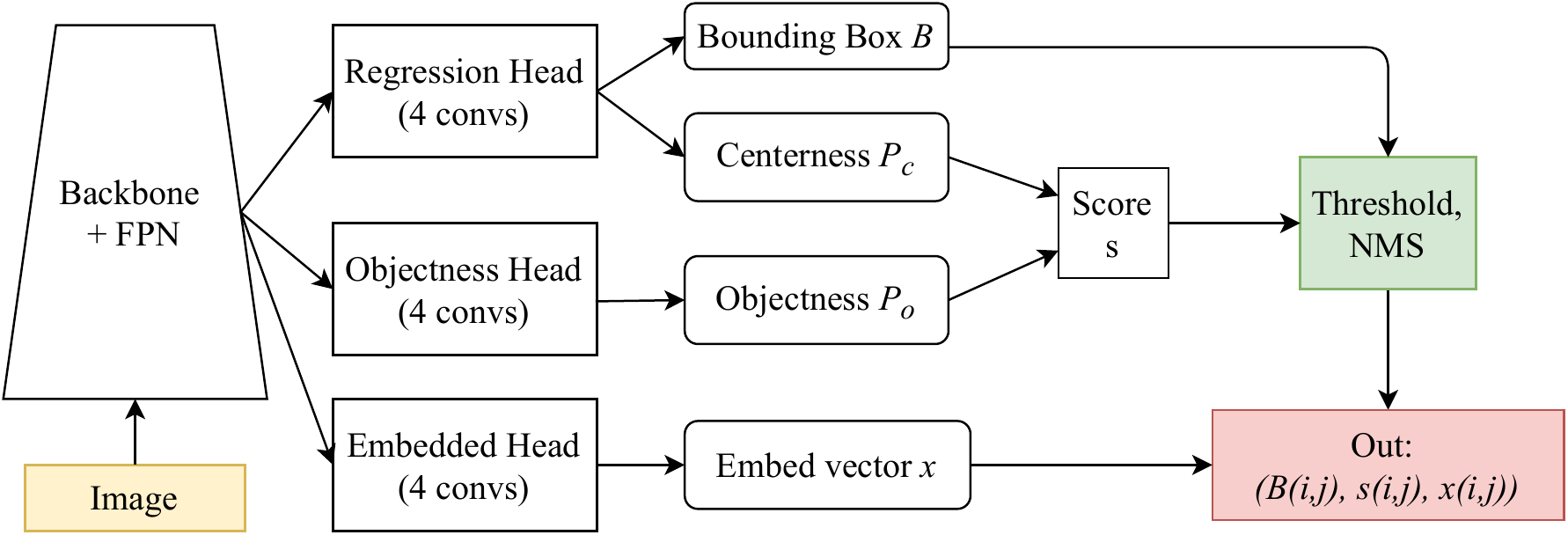}
 	\caption{Single Stage Common Object Detection Network (SSCOD).}
 	\label{fig:Network}
 \end{figure*}

Our proposed framework, Single Stage Common Object Detection Network (SSCOD), is illustrated in \Fig{Network}. The framework is built upon the standard Single Stage Object detection, such as Retina \cite{Lin2017} or FCOS \cite{Tian2019}. Specifically, the network includes a Backbone to extract features from an input image, and a Feature Pyramid Network (FPN) \cite{Lin2016} to fuse features from different scales. Features extracted from P3-P7 of FPN, i.e. with resolution $(H/2^3,W/2^3)$ to $(H/2^7,W/2^7)$, are passed to the detection head. 

We design the SSCOD head based on the Retina Head with ATSS \cite{Zhang2019c} sampling thanks to its efficiency, although other modules such as FCOS \cite{Tian2019} or Centerness \cite{Zhou2019}, \cite{Duan2019a} can be easily substituted. In particular, the detection head has 3 branches, namely Regression, Objectness, and Embedded Head, as illustrated in \Fig{Network}. The regression branch regresses to bounding box location $B=(x,y,w,h)$, and also predicts the object's center $p_c$, a.k.a centeredness \cite{Zhang2019c}. The objectness predicts the probability $p_o$ that a bounding box contains an object. This is similar to the classification branch in Retina, but in a class-agnostic manner, e.g only predict foreground vs. background. A bounding box centering at $(i,j)$ is considered as a valid object if its score
\begin{equation}
s(i,j) = p_o(i,j)p_c(i,j)
\end{equation}
is greater than a threshold.

To perform common class matching, we add an embedded branch to produce a representation vector $\x \in \Re^d$, where $d$ is the embedded dimension and $\x$ is normalized $\|\x\|=1$. Hence, a predicted box $b_i$ is represented by a tuple $b_i=(B_i,s_i,\x_i)$, and the similarity between two predicted bounding boxes $b_1$ and $b_2$ is measured by  
\begin{equation}
{\rm sim}(b_1,b_2)=s_1 s_2 \cos(\x_1,\x_2)=s_1 s_2 \x_1\Tr \x_2
\end{equation}

Following the ATSS sampling strategy \cite{Zhang2019c}, we only use a single scale, square anchor, for anchor setting. To keep it simple, similar to Retina or Mask-RCNN, each branch has 4 convolution layers kernel $3 \times 3$,  although other add-on blocks such as Deformable Conv \cite{zhu2019deformable}, Nonlocal Block \cite{wang2018non} can be easily added. To accommodate for small batch size, we use Convolution with Weight Standardized \cite{Qiao2019} followed by a Group Normalization (GN) \cite{wu2018group} and a ReLu activation. Each branch ends by a convolution layers kernel $3 \times 3$ without the normalization layer. For the objectness and centeredness branch, the features are passed through a Sigmoid layer.   

\subsection{Loss functions}\label{sec:Loss}
The Generalized IoU (GIoU) \cite{Rezatofighi2019} and the Cross-Entropy losses are utilized to train the bounding box regression and the centeredness branch. To train the objectness branch, we adopt an adaptive version of Focal Loss proposed by \cite{Weber2019}. For the embedded branch, we consider two types of loss functions for metric representation learning, namely class-wise and pair-wise losses.\\

\noindent \textbf{Class-wise losses}, such as Angular Loss \cite{Wang2017c}, SphereFace \cite{Liu2017b}, CosFace \cite{Wang2018b}, ArcFace \cite{Deng2019}, use a linear layer $W \in \Re^{d \times n}$ to map the embedded feature dimension to the number of classes $n$, followed by a softmax layer:
\begin{equation}\label{eq:l1}
L = -\frac{1}{N} \sum_{i=1}^{N} \log \frac{e^{W\Tr_{y_i}\x_i}}{\sum_{j=1}^n e^{W\Tr_{y_j}\x_i}}
\end{equation}
where $N$ is the batch size, and $W_j \in \Re^d$ denotes the j-th column of $W$, corresponding to class $y_j$. To enforce feature learning, the weight $W$ and features are normalized, e.g. $\|W_j\|=1$ $\|\x_i\|=1$, which leads to $W\Tr_j x_i = \|W_j\| \|\x_i\| \cos (\theta_j)= \cos (\theta_j)$. Hence, optimizing the loss function \eqn{l1} only depends on the angle between the feature and the weight, where $W_j$ can be associated as the center of class $y_i$. To smoothen the loss, the cosine value is often multiplied with a scale $s$ before computing softmax. 

Let $T(\theta_{y_i})$ and $N(\theta_j)$ be functions that modulate the angles between positive and negative samples respectively. In the simplest case (no manipulation), $T(\theta_{y_i})=\cos(\theta_{y_i})$, and $N(\theta_j)=\cos(\theta_j)$, and \eqn{l1} can be rewritten as:
\begin{equation}\label{eq:SMloss}
L = -\frac{1}{N} \sum_{i=1}^{N} \log \frac{e^{s T(\theta_{y_i})}}{e^{sT(\theta_{y_i})} + \sum_{j=1, j\neq y_i}^n e^{sN(\theta_{j})}}
\end{equation} 
Many approaches focus on modulating positive samples; for example,  in ArcFace Loss, a margin $m$ is added to the angle of positive samples:
\begin{equation}\label{eq:ArcFace}
T(\theta_{y_i})= \cos(\theta_{y_i}+m), \hspace{0.2in} N(\theta_j)=\cos(\theta_j)
\end{equation}
However, negative samples are also important. In metric representation learning, a negative sample can be classified as: (a) hard if $\theta_j < \theta_{y_i}$, (b) semi-hard if $\theta_{y_i} \leq \theta_j < \theta_{y_i} + m $, and (c) easy if $\theta_{y_i} + m \leq \theta_j$.
Curriculum Loss \cite{Huang2020}  further imposes a modulation to the negative samples:
\begin{equation}\label{eq:Curriculumn}
\begin{split}
T(\theta_{y_i}) &= \cos(\theta_{y_i}+m),\\
N(\theta_j) & = \begin{cases}
\cos(\theta_j) & \text{if } \theta_{y_i} + m \leq \theta_j\\
\cos(\theta_j)(t+\cos(\theta_j)) & \text{otherwise}
\end{cases}	
\end{split}
\end{equation}

That is, the weights of hard and semi-hard are adjusted during training by a modular $w=t + \cos(\theta_j)$. Here, $t$ is set to the average of positive cosine similarity
\begin{equation}\label{eq:t_avg}
t=\frac{\sum_i^N \cos(\theta_{y_i)}}{N}.
\end{equation}
At the beginning, $t\approx 0$, thus $w <1$ and the effect of hard negatives lessens, letting the model learn from the easy negative samples first. As the model begins to converge, it can detect negative samples better. Therefore, the number of easy negative samples increases, i.e. $\theta_{y_i} \rightarrow 0$, hence increasing the weight $t$, and switching the model's focus from easy to the hard negative samples. 

We adopt the Curriculum Loss to compare in our experiments. Furthermore, to deal with typical class-imbalance problem of object detection, we also impose a focal term:
\begin{equation}\label{eq:FoCur}
L = -\frac{1}{N} \sum_{i=1}^{N} (1 - p_i)^{\gamma(t)} \log (p_i), \\
\end{equation}
where 
\begin{equation}
p_i =  \frac{e^{s T(\theta_{y_i})}}{e^{sT(\theta_{y_i})} + \sum_{j=1, j\neq y_i}^n e^{sN(\theta_{j})}}
\end{equation} 
Inspired by Automated Focal Loss \cite{Weber2019}, we set $\gamma(t) = -\log(\max(t,10^{-5}))$. In practice, $t$ is computed through Exponential Moving Average of \eqn{t_avg}.\\  

\noindent \textbf{Pair-wise losses}, such as Max Margin Contrastive Loss \cite{Hadsell2006}, Triplet Loss \cite{Weinberger2009} \cite{Schroff2015}, \cite{Hermans2017}, Multi-class N-pair loss \cite{Sohn2016}, directly minimize the distances between different samples having same classes (positive pairs) and maximizes the distance between those of different labels (negative pairs). \\

For an anchor sample $\x_i$ in a data batch, we can find a set of positive pairs $U_i$ and negative pairs $V_i$. Let $N_i^+$ and $N_i^-$ as the size of $U_i$ and $V_i$ respectively, and $d_{ij}^+$ ($d_{ij}^-$) be the distance between two positive (negative) samples $\x_i$ and $\x_j$, where $d(\x_i,\x_j)=-\x_i\Tr \x_j$\footnote{The exact formula is $d(\x_i,\x_j)=1-\frac{\x_i\Tr \x_j}{\|\x_i\|\|\x_j\|}$, but we require $\|\x_i\|=\|\x_j\|=1$ and drop constant 1 for notation convenience since it does not affect the loss value.} for Cosine distance or $d(\x_i,\x_j)=\|\x_i-\x_j\|_2^2$ for Euclidean distance. When $N_i^+=1$ (or $N_i^-=1$ ), we denote $d_i^+$ (or $d_i^-$) as the only distance in the set. The general form of pairwise loss is:
\begin{equation}
L= \sum_{i}^N F ( \underset{j \in U_i}{d_{ij}^+}, \underset{k \in V_i}{d_{ik}^-}) 
\end{equation}

In Triplet Loss \cite{Hermans2017}, for example, $F = \max(0, m + \underset{j \in U_i}{\max }~d_{ij} - \underset{k \in V_i}{\min }~d_{ik})$, where $m$ is the margin. In  Multi-class N-pair loss \cite{Sohn2016}, the loss function is defined as: 
\begin{equation}\label{eq:NPairs}
\begin{split}
F &= \log(1 +  \underset{k \in V_i}{\sum}\exp(d_i^+ - d_{ik}^-)) \\
  &= -\log \frac{\exp(-d_i^+)}{\exp(-d_i^+) +  \underset{k \in V_i}{\sum} \exp(-d_{ik}^-)} 
\end{split}
\end{equation}
\eqn{NPairs} is also extended by using a temperature, as named NT-Xent by \cite{Chen2020}, and applied to general case where $N_i^+ \geq 1$, as named Supervised Contrastive loss by \cite{Khosla2020}:
\begin{equation}\label{eq:SupCon}
F = -\frac{1}{N_i^+}\underset{j \in U_i}{\sum} \log \frac{\exp(-d_{ij}^+/\tau)}{\exp(-d_{ij}^+/\tau) +  \underset{k \neq i}{\sum} \exp(-d_{ik}^-/\tau)} 
\end{equation}

Note that, the difference between \eqn{NPairs} and \eqn{SupCon} also lie in the denominator. In N-Pair loss \eqn{NPairs}, only the negative pairs are considered $(j \in V_i)$, while in \eqn{SupCon}, all positive and negative are used ($k\neq i$). This is because the NT-Xent and Supervised Contrastive loss are originally designed for transfer learning, aiming to learn general visual representation. In our case, since we aim for both class intra compactness and inter separability, we also consider the negative pairs only as in \eqn{NPairs}. 

Finally, let $s=1/\tau$ and using the Cosine distance, \eqn{SupCon} is rewritten as:
\begin{equation}\label{eq:SupConCos}
F = -\frac{1}{N_i^+}\underset{j \in U_i}{\sum} \log \frac{\exp(s\cos(\theta_{ij}))}{\exp(s\cos(\theta_{ij})) +  \underset{k \neq i}{\sum} \exp(s\cos(\theta_{ik}))} 
\end{equation}
Note that \eqn{SMloss} and \eqn{SupConCos} share the same structure. Therefore, we can adopt the modulation defined in \eqn{ArcFace} or \eqn{Curriculumn} to rewrite \eqn{SupConCos} as:
\begin{equation}\label{eq:ModSupCon}
F = -\frac{1}{N_i^+}\underset{j \in U_i}{\sum} \log \frac{e^{s T(\theta_{ij})}}{e^{s T(\theta_{ij})} +  \underset{k \in V_i}{\sum} e^{s N(\theta_{ik})}} 
\end{equation}

\begin{table*}[t]
	\vspace{1mm}
	\caption{Summary of Loss Functions for embedded matching. }
	\label{tab:SumLosses}
	\centering
	\resizebox{0.98\textwidth}{!}{%
		\begin{tabular}{|l|l|l|l|l|}
			\hline
			Loss                             & \begin{tabular}[c]{@{}l@{}}Focal Curriculum\\ (FocalCur)\end{tabular} & \begin{tabular}[c]{@{}l@{}}Arc Contrastive\\ (ArcCon)\end{tabular} & \begin{tabular}[c]{@{}l@{}}Arc Contrastive -Negative\\ (ArcCon-Neg)\end{tabular} & \begin{tabular}[c]{@{}l@{}}Curriculum Contrastive \\ (CurCon)\end{tabular} \\ \hline
			Formular Equ.                        &  \eqn{FoCur}            &       \eqn{ModSupCon}, $k\neq i$ as in \eqn{SupConCos}        &        \eqn{ModSupCon}, $k \in V_i$  as in \eqn{NPairs}                                                                         &   \eqn{ModSupCon}, $k\neq i$ as in \eqn{SupConCos}                                                                         \\ \hline
			$T(\theta), N(\theta)$ &           \eqn{Curriculumn}           &              \eqn{ArcFace}              &       \eqn{ArcFace}                           &              \eqn{Curriculumn}                                                              \\ \hline
	\end{tabular}}
\end{table*}

We name the former combination as \textbf{Arc Contrastive (ArcCon)} loss and the later as\textbf{ Curriculum Contrastive (CurCon)} loss. For CurCon Loss, we compute  $t=\frac{1}{N}\sum_{i=1}^N \underset{j \in U_i}{\min}(\cos{\theta_{ij}})$. 
\Tab{SumLosses} summarizes the loss functions proposed in this paper.

\section{\uppercase{Experiments}}
\subsection{Experimental Setup}

Our experiments are conducted on the popular PASCAL VOC \cite{Everingham2010} and the large-scale detection benchmarks COCO 2014 \cite{Lin2014a}. Following the common practice \cite{zhang2018single}, \cite{Zoph2019}, \cite{Zhang2019e}, we use both the trainval VOC 2007 and VOC 2012 for training (21.5k images and replicate 3 times), and evaluate on VOC2007 (5K images) test set. For the COCO dataset, we use the trainval35k split (115K images) for training and report the results on the minival split (5K images). Our code is implemented based on MMDetection opensource \cite{Chen2019d}.

\noindent \textbf{Training Details}: We conduct the experiments using the standard setup with backbone Resnet50 (R50) pre-trained from ImageNet. The stem convolution, the first stage, and Batch Norm layers of backbone R50 are frozen. For FPN and Head, we use Weight Standardized Convolution \cite{Qiao2019} and Group Norm \cite{wu2018group}. The model is trained with stochastic gradient descent (SGD) with momentum 0.9 and weight decay $10^{-4}$, with batch sizes (BS) equal 16. We train the models with 12 epochs for both VOC and COCO datasets.

We evaluate the effectiveness of both class-wise losses and pair-wise losses presented in \sec{Loss}. If a pair-wise loss is used, for each image in the batch, we also randomly sample another image that has at least one common object to form a valid pair. Hence the number of images in the batch is doubled. Therefore, we reduce the batch size a haft to make consistent training setting, but still keep the notation of batch size unchanged. In addition, we use auto policy V0 \cite{Zoph2019} for data augmentation. For the VOC dataset, the shorter side of the input images is resized to 600 pixels, while ensuring the longer side is smaller than 1000 pixels, and the aspect ratio is kept unchanged. In short, we denote it as resize to $(1000,600)$. For the COCO dataset, the images are set to a size of $(1333,800)$. All images after resizing are padded to be divisible by 32.    

We investigate different losses presented in \Tab{SumLosses}. To find appropriate parameters for the loss functions, we first train and evaluate the model on the validation set of VOC2007. We observe that ArcFace and CurricularFace are sensitive to parameter $s$, and $s=4, m= 0.5$ yields the best result. For pair-wise losses, $s\in [1,4], m=0.5$ yields quite equivalent results, and select $s=1, m=0.5$ as the default value for our experiments. 

\noindent \textbf{Inference and Evaluation Details}:
We first forward the input image through the network and obtain the predicted bounding boxes with a confidence score and embedded feature vector. We use the same post-processing parameters with RetinaNet (threshold 0.05 and NMS with maximum of 100 bounding boxes per image). For inference, two bounding boxes are considered belonging to a common class if their similarity score is greater than a threshold. 

For evaluation, we extract the top 100 pairs that have the highest matching scores from all possible matching pairs of two images. We evaluate the model using both Recall and Average Precision (AP) with the VOC evaluation style. Specifically, among the top 100 pairs, a predicted pair is true positive if it satisfies both conditions: (a) Each bounding box has $IoU>0.5$ with ground-truth boxes. (b) Their ground-truth boxes have the same object categories. Otherwise, it is a false positive. To generate ground truth, although arbitrary number of pairs can be used, we follow \cite{Jiang2019a} to randomly sample $p=6$ valid pairs for each image in the validation dataset, where the random seed is set to 0 for reproducibility.\\

\noindent \textbf{Baseline Model} The easiest solution for common object detection is to use the standard object detection approach, then a common pair can be estimated by:
\begin{itemize}
	\item Hard matching (HM): Object category of each bounding box can be inferred as the class having the highest probability. Two bounding boxes form a valid pair if they predict the same class.
	\item Soft matching (SM): The probability score is used as a description vector, and matching score can be computed as their cosine similarity. The inference hence follows the same setup of our solution.  
\end{itemize}

\subsection{Pascal VOC}
The learning rate is linearly increased using warm-up strategy to $0.5e^{-2}$ in the first 300 iterations, and then gradually reduced by the cosine annealing to $0.5e^{-4}$. \\
\begin{table*}[!t]
	\vspace{2mm}
	\captionsetup{width=0.9\textwidth}
	\caption{Comparison between the baseline model and SSCOD model for Experiment Type 1 on VOC dataset . \textbf{Bold} and \textbf{\textit{Both Italic}} represent the best results of baseline and SSCOD, respectively.}
	\label{tab:VOC-exp1}
	\centering
	\begin{tabular}{|c|c|c|c|c|c|c|}
		\hline
		\multirow{2}{*}{\begin{tabular}[c]{@{}c@{}}Eval.\\ Metrics\end{tabular}} & \multicolumn{2}{c|}{Baseline}    & \multicolumn{4}{c|}{SSCOD}                          \\ \cline{2-7} 
		& HM              & SM              & FocalCur & ArcCon & ArcCon-Neg & CurCon \\ \hline
		Recall                                                                   & 0.7922          & \textbf{0.7976} & 0.7208     & 0.7771 & 0.6356     & \textbf{\textit{0.8090}} \\ \hline
		AP                                                                       & \textbf{0.6052} & 0.5746          & 0.5986     & 0.5832 & 0.5326     & \textbf{\textit{0.6141}} \\ \hline
	\end{tabular}
\end{table*}

\begin{table*}[!t]
	\captionsetup{width=0.9\textwidth}
	\caption{Comparison between the baseline model and SSCOD model for Experiment Type 2 on the VOC dataset. \textbf{Bold} and \textbf{\textit{Both Italic}} represent the best results of baseline and SSCOD, respectively.}
	\label{tab:VOC-exp2}
	\centering
	\begin{tabular}{|c|c|c|c|c|c|c|c|}
		\hline
		\multirow{2}{*}{Type} & \multirow{2}{*}{\begin{tabular}[c]{@{}c@{}}Eval.\\ Metrics\end{tabular}} & \multicolumn{2}{c|}{Baseline} & \multicolumn{4}{c|}{SSCOD}                                  \\ \cline{3-8} 
		&                                                                          & HM                 & SM        & FocalCur & ArcCon          & ArcCon-Neg & CurCon          \\ \hline
		\multirow{2}{*}{EONC} & Recall                                                                   & \textbf{0.8128}             & 0.8123    & 0.8078     & 0.8232          & 0.7752     & \textbf{\textit{0.8385}} \\ \cline{2-8} 
		& AP                                                                       & \textbf{0.6278}    & 0.6001    & 0.5443     & 0.5505          & 0.4798     & \textbf{\textit{0.5857}}          \\ \hline
		\multirow{2}{*}{ONC}  & Recall                                                                   & 0.0699             & 0.0850    & 0.5925     & 0.6211          & 0.5936     & \textbf{\textit{0.6267}} \\ \cline{2-8} 
		& AP                                                                       & 0.0012             & 0.0014    & 0.1906     & \textbf{\textit{0.2765}} & 0.2510     & 0.2663          \\ \hline
	\end{tabular}
\end{table*}

\begin{table*}[!t]
	\captionsetup{width=0.9\textwidth}
	\caption{Comparison between the baseline model and SSCOD model for Experiment Type 3 on the VOC dataset. \textbf{Bold} and \textbf{\textit{Both Italic}} represent the best results of baseline and SSCOD, respectively.}
	\label{tab:VOC-exp3}
	\centering
	\begin{tabular}{|c|c|c|c|c|c|c|c|}
		\hline
		\multirow{2}{*}{Type} & \multirow{2}{*}{\begin{tabular}[c]{@{}c@{}}Eval.\\ Metrics\end{tabular}} & \multicolumn{2}{c|}{Baseline} & \multicolumn{4}{c|}{SSCOD}                                  \\ \cline{3-8} 
		&                                                                          & HM                 & SM        & FocalCur & ArcCon          & ArcCon-Neg & CurCon          \\ \hline
		\multirow{2}{*}{EONC} & Recall                                             & \textbf{0.8128}    & 0.8123    & 0.7347     & 0.7509    & 0.4103     & \textbf{\textit{0.8105}} \\ \cline{2-8} 
		& AP                                                                       & \textbf{0.6278}    & 0.6001    & 0.4262     & 0.4595    & 0.2572     & \textbf{\textit{0.5593}}          \\ \hline
		\multirow{2}{*}{ONC}   & Recall                                             & 0.0699    & \textbf{0.0850 }   & 0.3563     & \textbf{\textit{0.4060}}    & 0.3418     & 0.3810 \\ \cline{2-8} 
		& AP                                                                       & 0.0012    & \textbf{0.0014}    & 0.066      & \textbf{\textit{0.1391}}    & 0.0567     & 0.1212          \\ \hline
	\end{tabular}
\end{table*}

\noindent \textbf{Experiment Type 1}: First, to verify if our SSCOD degrades the accuracy for known categories, we train the standard model ATSS with R50 backbone by common setting, and infer the Hard-matching (HM) and Soft-Matching (SM) baseline. In this experiment, our baseline model without test-time augmentation or complex structure achieves 0.774 mAP, which is similar or higher than several benchmark results \cite{Ren2015}, \cite{Liu2016} \cite{Fu2017}. This validates that our model can be a strong baseline. Similarly, we train the proposed SSCOD model with different losses. The experiments are conducted using all samples of 20 classes, and the result is presented in \Tab{VOC-exp1}.

\noindent As seen from \Tab{VOC-exp1}, results of SSCOD are asymptotic to those of the baseline. Specifically, SSCOD trained with FocalCur loss achieves 0.5986 AP, which is closely matched to the HM baseline of 0.6052 AP and higher than SM baseline of 0.5746 AP. The CurCon loss obtains the best result in this case, which is $0.6141 AP$ and higher than both baselines. The ArcCon loss yields slightly worse result than the HM baseline, but still better than SM baseline.\\
Note that, images in VOC dataset often has only one or two classes. Hence, the performance of baseline for VOC is quite predictable from its mAP, since the precision when predicting a pair is conditional on the precision of each box component, e.g. $mAP^2=0.774^2 \approx 0.6$. The hard matching yields higher accuracy than soft-matching because it can suppress the noise better through the post processing. Nevertheless, our SSCOD using cosine similarity can achieve comparable or better results. Surprisingly, ArcCon-Neg yields the worst result. This is possibly due to insufficient number of negative pairs, which is even more severe due to small batch size used in object detection.  \\

\noindent \textbf{Experiment Type 2}: Second, to evaluate the ability to detect common pairs of unseen categories, we remove the five classes from the training set. For simplicity and reproducibility, we chose the last five classes, namely: \textit{potted plant, sheep, sofa, train} and \textit{tv monitor}, and repeat similar experiments on the truncated training set. 
Different from the known-category case, the unseen case depends on both the localization ability of the objectness branch and the matching accuracy of embedded features. Hence, to independently evaluate the matching module, we train the objectness branch using all samples from the dataset. We emphasize that this is solely to learn foreground and background concept, hence no object category information is used. Follow \cite{Jiang2019a}, we denote unseen classes as Only Novel Categories (ONC) and seen classes as Excluded ONC (EONC), and the experiment results are shown in \Tab{VOC-exp2}. 

As seen from \Tab{VOC-exp2}, although the baseline model yields the highest results for EONC test, it can't detect novel classes in ONC test. This is expected due the standard object detection is only designed for close-set condition. In contrast, our SSCOD is still able to match objects from unseen classes. Specifically, ArcCon loss yields the best result for ONC test, $0.2765 AP$. On the other hand, CurCon loss gets a better balance between EONC and ONC tests, $0.5857 AP$, and $0.2663 AP$ for objects of seen and unseen categories respectively. FocalCur loss yields worse results for ONC test in this case. \\

\noindent \textbf{Experiment Type 3}: Finally, we evaluate the models in the most restricted case, where both objectness and embedded branches are trained with samples from only 15 seen classes. In practice, since one image may contain several objects, simply dropping samples of unknown classes will force the network treating them as background, thus harm the network generalization. To overcome this problem, we simply set the weight for samples of unknown classes equal 0 during computing the loss for objectness branch and regression branches. Note that, this setting does not restrict real application because in practice all samples are used to train the network, and unlabeled objects are often not interesting. For the embedded branch, these samples are totally ignored. The highest results of each experiment are presented in \Tab{VOC-exp3}.

As seen from \Tab{VOC-exp3}, ArcCon and CurCon perform best in this case, while FocalCur and ArcCon-Neg perform worst. The CurCon loss still yields the best balance between EONC and ONC test, $0.5593 AP$ and $0.1212 AP$ for seen and unseen categories respectively. However, different from previous experiments, we observe the trade-off between EONC and ONC's performance. 

\begin{figure}[!h]
	\centering
	\includegraphics[width=0.98\columnwidth]{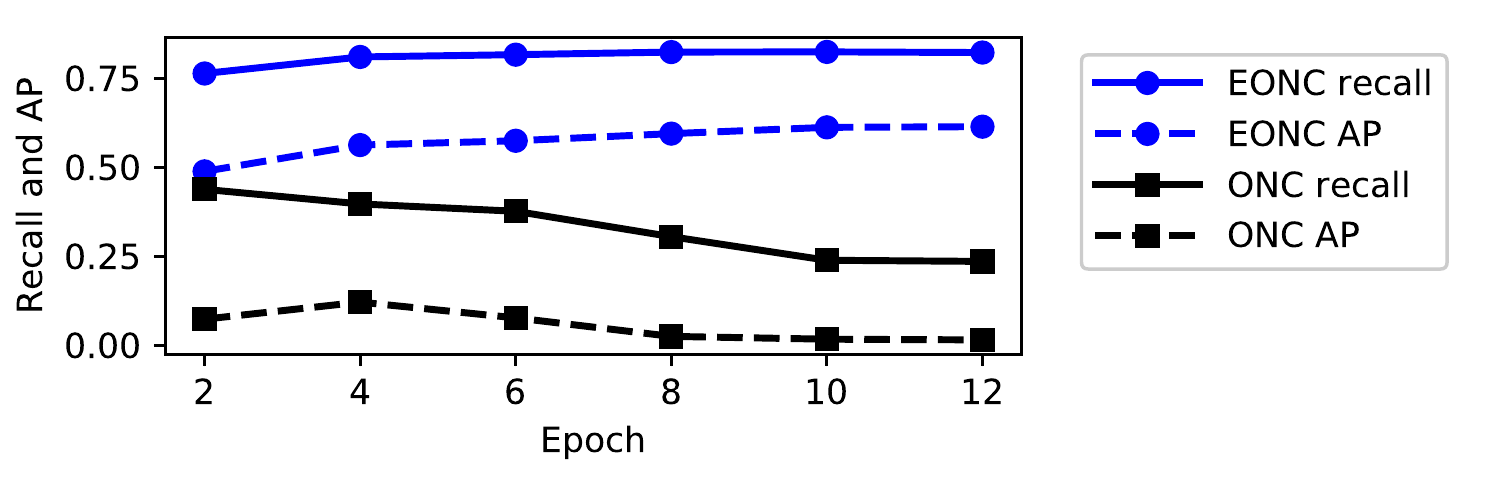}
	\caption{Trade-off between EONC and ONC test in Experiment Type 3 using CurCon Loss}
	\label{fig:exp3}
\end{figure}

As illustrated in \fig{exp3}, the longer training, the better EONC test but the worse ONC test. This is because the objectness and regression branches start to overfit to the seen classes, hence decrease the localization ability for unseen classes. This effect does not happen in Type 2 experiments.

\subsection{COCO 2014}
We conduct similar experiments on the COCO dataset for type 1 and 2 as done in VOC.  

\noindent \textbf{Experiment Type 1}: The learning rate is linearly increased using warm-up strategy to $1e^{-2}$ in the first 500 iterations, and then gradually reduced by the cosine annealing to $1e^{-4}$. Since ArcCur-Neg loss does not yield comparable results with other, we exclude it for COCO experiments. For the baseline, we use the checkpoint \footnote{https://github.com/open-mmlab/mmdetection/tree/master/configs/atss} provided by MMDetection, which has 39.4 box mAP.  
\hskip-0.1cm
\begin{table*}
	\vspace{2mm}
	\captionsetup{width=0.95\textwidth}
	\caption{Comparison between the baseline model and SSCOD model for Experiment Type 1 on the COCO dataset. \textbf{Bold} and \textbf{\textit{Both Italic}} represent the best results of baseline and SSCOD, respectively.}
	\label{tab:CoCo-exp1}
	\centering
	\begin{tabular}{|l|l|l|l|l|l|}
		\hline
		\multirow{2}{*}{\begin{tabular}[c]{@{}l@{}}Eval \\ Metrics\end{tabular}} & \multicolumn{2}{c|}{Baseline} & \multicolumn{3}{c|}{SSCOD} \\ \cline{2-6} 
		& HM            & SM            & FocalCur & ArcCur & CurCon \\ \hline
		Recall                                                                   & \textbf{0.5128}        & 0.5102        & 0.5101   & 0.5473 & \textbf{\textit{0.5862}} \\ \hline
		AP                                                                       & \textbf{0.3688}        & 0.3515        & 0.3615   & 0.3160 & \textbf{\textit{0.3811}} \\ \hline
	\end{tabular}
\end{table*}

\begin{table*}[!t]
	\captionsetup{width=0.95\textwidth}
	\caption{Comparison between the baseline model and SSCOD model for Experiment Type 2 on the COCO dataset. \textbf{Bold} and \textbf{\textit{Both Italic}} represent the best results of baseline and SSCOD, respectively.}
	\label{tab:CoCo_exp2}
	\centering
	\begin{tabular}{|l|l|l|l|l|l|l|l|}
		\hline
		\multirow{3}{*}{Type} & \multirow{2}{*}{\begin{tabular}[c]{@{}l@{}}Eval \\ Metrics\end{tabular}} & \multicolumn{3}{c|}{Case A}             & \multicolumn{3}{c|}{Case B}             \\ \cline{3-8} 
		&                                                                          & \multicolumn{2}{c|}{Baseline} & SSCOD   & \multicolumn{2}{c|}{Baseline} & SSCOD   \\ \cline{2-8} 
		& Metrics                                                                  & HM            & SM            & CurCon  & HM            & SM            & CurCon  \\ \hline
		\multirow{2}{*}{EONC} & Recall                                                                   & \textbf{0.5141}        & 0.5075        & \textbf{\textit{0.5862}}  & \textbf{0.5247}        & 0.5206        & \textbf{\textit{0.6201}}  \\ \cline{2-8} 
		& AP                                                                       & \textbf{0.3676}        & 0.3501        & \textbf{\textit{0.3811}}  & \textbf{0.3781}        & 0.3603        & \textbf{\textit{0.4074}}  \\ \hline
		\multirow{2}{*}{ONC}  & Recall                                                                   & 0.0545             & \textbf{0.0595}             & \textbf{\textit{0.4202}}  & \textbf{0.0739}             & 0.0540             & \textbf{\textit{0.4587}}  \\ \cline{2-8} 
		& AP                                                                       & 0.0003             & 0.0003             &\textbf{\textit{0.0643}}  & 0.0003             & 0.0003             & \textbf{\textit{0.1213}}  \\ \hline
	\end{tabular}
\end{table*}

The results are shown in \Tab{CoCo-exp1}. In this case, CurCon loss yields the best performance and surpasses the base-line with a large margin for both recall and AP, 0.5862 and 0.3811 respectively. FocalCur loss also yields closely matching results with the baseline. These results are consistent with VOC's results, hence confirm the effectiveness of our proposed approach. 

\noindent \textbf{Experiment Type 2}: Unlike VOC, COCO dataset has many fine-granded classes in each meta-classes, namely: \textit{person, vehicle, outdoor, animal, accessory, sports, kitchen, food, furniture, electronics, appliance, and indoor.}  Therefore, selecting seen and unseen categories for training and evaluation can have a significant effect on the AP score. For example, if both car and bus are in unseen classes, their high score matching is reasonable but false-positive. Therefore, we conduct two experiments:
\begin{itemize}
	\item Case A: Follow \cite{Jiang2019a}, we select 30 classes to train the model. However, since they do not specify the class names, reproducing is hard. Here, we choose the training classes as: \textit{person, bicycle, car, airplane, boat, fire hydrant, stop sign, dog, horse, elephant, umbrella, handbag, snowboard, sports ball, baseball bat, skateboard, bottle, fork, bowl, apple, carrot, cake, chair, toilet, laptop, cell phone, microwave, sink, book, and hair drier}.
	\item Case B: We split 75\% classes for training, and 25\% other for testing. Concretely, 20 unseen classes are: \textit{motorcycle, bus, cat, horse, sheep, backpack, tie, skis, sports ball, surfboard, tennis racket, cup, banana, hot dog, pizza, donuts, remote, toaster, clock, teddy bear}.
\end{itemize}
We report the results using only CurCon loss in \Tab{CoCo_exp2}, since we found it yields the best results in all previous experiments. For EONC test, SSCOD yields higher results than the baseline with a large margin, $0.3811AP$ and $0.4074AP$ for case A and B respectively. As previously mentioned, the baseline can not work for ONC test. For Case A, the performance is poor due to the shortage of training samples. For Case B, the results are better, e.g. $0.1213AP$, due to a larger number of training samples.

\subsection{Discussion}
\subsubsection{Comparison of Loss functions}

Although the class-wise and pairwise losses have been used for Face-ID or unsupervised learning, this work adopts them for (unseen) object detection. In our experiments, the pairwise losses are more effective than the classwise loss. We hypothesize that this is because in classwise loss, the number of contrastive pairs in the denominator is limited to the number of known classes (the centroid of each class) from the train set. In contrast, the number of pairs in the denominator of pairwise losses is essentially all possible object pair combinations, governed by the number of bounding boxes in a minibatch. This helps increase the interaction between the sample pairs, and especially useful for the case of unknown classes detection. Our results are in-line with recent research of contrastive learning \cite{Chen2020} \cite{he2020momentum}, who also finds the importance of using a large training batch size to increase the number of negative pairs for good performance. 

Our proposed Curriculum Contrastive Loss performs best in most of experiments, since it unifies both approaches by adding the adaptive angular margins to the contrastive loss formulation.

\subsubsection{Comparison to previous works}
\begin{table*}[]
	\vspace{2mm}
	\captionsetup{width=0.95\textwidth}
	\caption{Comparison between the proposed method and previous works on the VOC dataset. \textbf{Bold} represents the best results.}
	\label{tab:prev_works_voc}
	\centering
	\begin{tabular}{|c|c|c|c|c|c|c|c|}
		\hline
		\multirow{2}{*}{Type} & \multicolumn{4}{c|}{Ours}                                        & \multicolumn{3}{c|}{Jiang et al. 2019}    \\ \cline{2-8} 
		& FocalCur        & ArcCon          & ArcCon-Neg & CurCon          & Best Baseline & Siamese & Relation \\ \hline
		EONC                  & 0.5443      & 0.5505          & 0.4798     & \textbf{0.5857} & 0.4481        & 0.3269  & 0.3774   \\ \hline
		ONC                   & 0.1906          & \textbf{0.2765} & 0.251      & 0.2663          & 0.1638        & 0.2187  & 0.2535   \\ \hline
	\end{tabular}
\end{table*}

\begin{table*}[]
	\captionsetup{width=0.95\textwidth}
	\caption{Comparison between the proposed method and previous works on the COCO dataset. \textbf{Bold} represents the best results.}
	\label{tab:prev_works_coco}
	\centering
	\begin{tabular}{|c|c|c|c|c|}
		\hline
		\multirow{2}{*}{Type} & Ours            & \multicolumn{3}{c|}{Jiang et al. 2019}           \\ \cline{2-5} 
		& CurCon          & Best Baseline & Siamese & Relation        \\ \hline
		EONC                  & \textbf{0.3811} & 0.2107        & 0.141   & 0.1773          \\ \hline
		ONC                   & 0.0643          & 0.1247        & 0.1398  & \textbf{0.1824} \\ \hline
	\end{tabular}
\end{table*}

To compare with previous works, we attempt to reproduce the results of \cite{Jiang2019a}. However, this is challenging, due to missing information. Specifically, in our reproducing attempt, their proposed Siamese Network was unstable during training, e.g. when $sim(p_1,p_2)<0$ in their Equ (7), the loss is NaN. Also for the Siamese network, sampling strategy is critical for convergence but unmentioned. Similarly, in their proposed Relation Matching network, $concat(f_1,f_2)$ and $concat(f_2,f_1)$ can yield different results. Furthermore, missing information such as how to select 20 images pair for training, how to select seen/unseen classes, and setup optimization makes the reproducing process very difficult. Hence, we use their reported results for direct comparison, albeit possibly different settings.

As seen from \tab{prev_works_voc} and \tab{prev_works_coco}, their EONC results are always worse then the baseline, while ONC is better. This is questionable, since the model is trained on seen classes, it should perform better for EONC case. This is contrast with our results. In the Type 1 experiments, our SSCOD always performs better the baseline, which also has much higher AP than the Faster-RCNN baseline used by \cite{Jiang2019a}. This confirms the effectiveness of our proposed methods.

For ONC test, our results on VOC is still higher than the Siamese and Relation Network methods reported by \cite{Jiang2019a}. In addition, our proposed approach does not require complicated training setup, offline sampling mechanism or extra matching modulation, therefore can serve as a good baseline. In contrast, our proposed method perform worse on COCO dataset, which is likely due to missing detection for unknown objects.  

However, there are still ample room to improve the results. As showed in \tab{VOC-exp3} and \fig{exp3}, the network can be easily trained to optimize the performance on seen classes, but this will reduce the ability to generalize for unseen objects. This problem can be partially alleviated by training on larger and more diverse dataset. In addition, we can also treat it as positive-unlabeled problem \cite{Yang2020} to reduce the effect of missing labels. Currently, the objectness, the regression and the embedded branches are trained independently, and this can be insufficient. Adding an attention mechanism from embedded features to the objectness can also enhance the results. We leave the discussion above for future work. 
\subsubsection{Result Visualization}
Due to space limitations, visualization of predicted results and the pretrained model to generate predictions can be found at  \href{https://anonymous.4open.science/r/10a4cbab-143c-4fe8-a246-af1ccdcb578b/}{(URL)}.

\section{\uppercase{Conclusion}}
This paper proposes a solution for common object detection, which aims to detect pairs of objects from similar categories in a set of images. While this is an interesting problem, there are many challenges, such as the ability to work on both closed-set and open-set conditions, and for multiple objects. Our solution is built upon single-stage object detection thanks to its efficiency. To matching objects of the same category, we add an embedded branch to the network to generate representation features. Several loss functions to train the embedded branch are investigated. The proposed Curriculum Contrastive loss, which combines contrastive learning and angular margin losses, gives the best performance. The experiments on both VOC and COCO dataset demonstrate that our approach yields higher accuracy than the base-line of standard object detection for both seen and unseen categories. We hope this work can serve as a strong baseline for future research of Common Object Detection.    

\bibliographystyle{apalike}
{\small
	\bibliography{CyberCore-Codet_paper}}

\begin{thebibliography}{}

\bibitem[Bao et~al., 2012]{Bao2012}
Bao, S.~Y., Xiang, Y., and Savarese, S. (2012).
\newblock {Object co-detection}.
\newblock {\em Lecture Notes in Computer Science}, 7572 LNCS(PART 1):86--101.

\bibitem[Cai and Vasconcelos, 2017]{Cai2018b}
Cai, Z. and Vasconcelos, N. (2017).
\newblock {Cascade R-CNN: Delving into High Quality Object Detection}.
\newblock {\em Proceedings of the IEEE Computer Society Conference on Computer
  Vision and Pattern Recognition}, pages 6154--6162.

\bibitem[Cai and Vasconcelos, 2019]{Cai2019}
Cai, Z. and Vasconcelos, N. (2019).
\newblock Cascade r-cnn: high quality object detection and instance
  segmentation.
\newblock {\em IEEE Transactions on Pattern Analysis and Machine Intelligence}.

\bibitem[Chen et~al., 2019a]{Chen2019b}
Chen, H., Huang, Y., and Nakayama, H. (2019a).
\newblock {Semantic Aware Attention Based Deep Object Co-segmentation}.
\newblock {\em Lecture Notes in Computer Science}, 11364 LNCS:435--450.

\bibitem[Chen et~al., 2019b]{Chen2019d}
Chen, K., Wang, J., Pang, J., Cao, Y., Xiong, Y., Li, X., Sun, S., Feng, W.,
  Liu, Z., Xu, J., Zhang, Z., Cheng, D., Zhu, C., Cheng, T., Zhao, Q., Li, B.,
  Lu, X., Zhu, R., Wu, Y., Dai, J., Wang, J., Shi, J., Ouyang, W., Loy, C.~C.,
  and Lin, D. (2019b).
\newblock {MMDetection: Open MMLab Detection Toolbox and Benchmark}.
\newblock {\em arXiv preprint arXiv:1906.07155}.

\bibitem[Chen et~al., 2020]{Chen2020}
Chen, T., Kornblith, S., Norouzi, M., and Hinton, G. (2020).
\newblock {A Simple Framework for Contrastive Learning of Visual
  Representations}.
\newblock {\em arXiv preprint arXiv:2002.05709}.

\bibitem[Deng et~al., 2019]{Deng2019}
Deng, J., Guo, J., Xue, N., and Zafeiriou, S. (2019).
\newblock {ArcFace: Additive angular margin loss for deep face recognition}.
\newblock {\em Proceedings of the IEEE Computer Society Conference on Computer
  Vision and Pattern Recognition}, 2019-June:4685--4694.

\bibitem[Duan et~al., 2019]{Duan2019a}
Duan, K., Bai, S., Xie, L., Qi, H., Huang, Q., and Tian, Q. (2019).
\newblock Centernet: Keypoint triplets for object detection.
\newblock In {\em Proceedings of the IEEE International Conference on Computer
  Vision}, pages 6569--6578.

\bibitem[Everingham et~al., 2010]{Everingham2010}
Everingham, M., {Van Gool}, L., Williams, C.~K., Winn, J., and Zisserman, A.
  (2010).
\newblock {The pascal visual object classes (VOC) challenge}.
\newblock {\em International Journal of Computer Vision}, 88(2):303--338.

\bibitem[Fu et~al., 2017]{Fu2017}
Fu, C.-y., Liu, W., Ranga, A., Tyagi, A., and Berg, A.~C. (2017).
\newblock {DSSD : Deconvolutional Single Shot Detector}.
\newblock {\em arXiv preprint arXiv:1701.06659}.

\bibitem[Gidaris and Komodakis, 2016]{Gidaris2016}
Gidaris, S. and Komodakis, N. (2016).
\newblock {Attend Refine Repeat : Active Box Proposal}.
\newblock {\em arXiv preprint arXiv:1606.04446v1}.

\bibitem[Girshick, 2015]{Girshick2015}
Girshick, R. (2015).
\newblock Fast r-cnn.
\newblock In {\em Proceedings of the IEEE international conference on computer
  vision}, pages 1440--1448.

\bibitem[Girshick et~al., 2014]{Girshick2014}
Girshick, R., Donahue, J., Darrell, T., and Malik, J. (2014).
\newblock {Rich feature hierarchies for accurate object detection and semantic
  segmentation}.
\newblock {\em Proceedings of the IEEE Computer Society Conference on Computer
  Vision and Pattern Recognition}, pages 580--587.

\bibitem[Guo et~al., 2013]{Guo2013a}
Guo, X., Liu, D., Jou, B., Zhu, M., Cai, A., and Chang, S.~F. (2013).
\newblock {Robust object co-detection}.
\newblock {\em Proceedings of the IEEE Computer Society Conference on Computer
  Vision and Pattern Recognition}, pages 3206--3213.

\bibitem[Hadsell et~al., 2006]{Hadsell2006}
Hadsell, R., Chopra, S., and LeCun, Y. (2006).
\newblock {Dimensionality reduction by learning an invariant mapping}.
\newblock {\em Proceedings of the IEEE Computer Society Conference on Computer
  Vision and Pattern Recognition}, 2:1735--1742.

\bibitem[He et~al., 2020]{he2020momentum}
He, K., Fan, H., Wu, Y., Xie, S., and Girshick, R. (2020).
\newblock Momentum contrast for unsupervised visual representation learning.
\newblock In {\em Proceedings of the IEEE/CVF Conference on Computer Vision and
  Pattern Recognition}, pages 9729--9738.

\bibitem[He et~al., 2017]{He2017a}
He, K., Gkioxari, G., Dollar, P., and Girshick, R. (2017).
\newblock {Mask R-CNN}.
\newblock {\em Proceedings of the IEEE International Conference on Computer
  Vision}, 2017-Octob:2980--2988.

\bibitem[Hermans et~al., 2017]{Hermans2017}
Hermans, A., Beyer, L., and Leibe, B. (2017).
\newblock {In Defense of the Triplet Loss for Person Re-Identification}.
\newblock {\em arXiv preprint arXiv:1703.07737}.

\bibitem[Huang et~al., 2020]{Huang2020}
Huang, Y., Wang, Y., Tai, Y., Liu, X., Shen, P., Li, S., Li, J., and Huang, F.
  (2020).
\newblock Curricularface: adaptive curriculum learning loss for deep face
  recognition.
\newblock In {\em Proceedings of the IEEE/CVF Conference on Computer Vision and
  Pattern Recognition}, pages 5901--5910.

\bibitem[Jiang et~al., 2019]{Jiang2019a}
Jiang, S., Liang, S., Chen, C., Zhu, Y., and Li, X. (2019).
\newblock {Class Agnostic Image Common Object Detection}.
\newblock {\em IEEE Transactions on Image Processing}, 28(6):2836--2846.

\bibitem[Joulin et~al., 2010]{Joulin2010}
Joulin, A., Bach, F., and Ponce, J. (2010).
\newblock {Discriminative clustering for image co-segmentation}.
\newblock {\em Proceedings of the IEEE Computer Society Conference on Computer
  Vision and Pattern Recognition}, pages 1943--1950.

\bibitem[Khosla et~al., 2020]{Khosla2020}
Khosla, P., Teterwak, P., Wang, C., Sarna, A., Tian, Y., Isola, P., Maschinot,
  A., Liu, C., and Krishnan, D. (2020).
\newblock {Supervised Contrastive Learning}.
\newblock {\em arXiv preprint arXiv:2004.11362}, pages 1--18.

\bibitem[Law and Deng, 2018]{Law2018}
Law, H. and Deng, J. (2018).
\newblock {Cornernet: Detecting objects as paired keypoints}.
\newblock {\em Lecture Notes in Computer Science}, 11218 LNCS:765--781.

\bibitem[Le et~al., 2017]{Le2017}
Le, H., Yu, C.~P., Zelinsky, G., and Samaras, D. (2017).
\newblock {Co-localization with Category-Consistent Features and Geodesic
  Distance Propagation}.
\newblock {\em Proceedings - 2017 IEEE International Conference on Computer
  Vision Workshops, ICCVW 2017}, 2018-Janua:1103--1112.

\bibitem[Li et~al., 2019a]{Li2019d}
Li, W., {Hosseini Jafari}, O., and Rother, C. (2019a).
\newblock {Deep Object Co-segmentation}.
\newblock In {\em Lecture Notes in Computer Science}, volume 11363 LNCS, pages
  638--653.

\bibitem[Li et~al., 2019b]{Li2019}
Li, W., Jafari, H., and Rother, C. (2019b).
\newblock {Localizing Common Objects Using Common Component Activation Map}.
\newblock pages 28--31.

\bibitem[Lin et~al., 2017a]{Lin2016}
Lin, T.-Y., Doll{\'a}r, P., Girshick, R., He, K., Hariharan, B., and Belongie,
  S. (2017a).
\newblock Feature pyramid networks for object detection.
\newblock In {\em Proceedings of the IEEE conference on computer vision and
  pattern recognition}, pages 2117--2125.

\bibitem[Lin et~al., 2017b]{Lin2017}
Lin, T.~Y., Goyal, P., Girshick, R., He, K., and Dollar, P. (2017b).
\newblock {Focal Loss for Dense Object Detection}.
\newblock {\em IEEE Transactions on Pattern Analysis and Machine Intelligence},
  42(2):318--327.

\bibitem[Lin et~al., 2014]{Lin2014a}
Lin, T.~Y., Maire, M., Belongie, S., Hays, J., Perona, P., Ramanan, D.,
  Doll{\'{a}}r, P., and Zitnick, C.~L. (2014).
\newblock {Microsoft COCO: Common objects in context}.
\newblock In {\em Lecture Notes in Computer Science}, volume 8693 LNCS, pages
  740--755.

\bibitem[Liu et~al., 2016]{Liu2016}
Liu, W., Anguelov, D., Erhan, D., Szegedy, C., Reed, S., Fu, C.~Y., and Berg,
  A.~C. (2016).
\newblock {SSD: Single shot multibox detector}.
\newblock {\em Lecture Notes in Computer Science}, 9905 LNCS:21--37.

\bibitem[Liu et~al., 2017]{Liu2017b}
Liu, W., Wen, Y., Yu, Z., Li, M., Raj, B., and Song, L. (2017).
\newblock {SphereFace: Deep hypersphere embedding for face recognition}.
\newblock {\em Proceedings - 30th IEEE Conference on Computer Vision and
  Pattern Recognition, CVPR 2017}, 2017-Janua:6738--6746.

\bibitem[Merdassi et~al., 2019]{Merdassi2019}
Merdassi, H., Barhoumi, W., and Zagrouba, E. (2019).
\newblock {A Comprehensive Overview of Relevant Methods of Image
  Cosegmentation}.
\newblock {\em Expert Systems with Applications}, 140:112901.

\bibitem[Qiao et~al., 2019]{Qiao2019}
Qiao, S., Wang, H., Liu, C., Shen, W., and Yuille, A. (2019).
\newblock {Weight Standardization}.
\newblock {\em arXiv preprint arXiv:1903.10520}.

\bibitem[Quan et~al., 2016]{7780450}
Quan, R., Han, J., Zhang, D., and Nie, F. (2016).
\newblock {Object Co-segmentation via Graph Optimized-Flexible Manifold
  Ranking}.
\newblock In {\em 2016 IEEE Conference on Computer Vision and Pattern
  Recognition (CVPR)}, pages 687--695.

\bibitem[Redmon and Farhadi, 2017]{Redmon2017}
Redmon, J. and Farhadi, A. (2017).
\newblock {YOLO9000: Better, faster, stronger}.
\newblock {\em Proceedings - 30th IEEE Conference on Computer Vision and
  Pattern Recognition, CVPR 2017}, 2017-Janua:6517--6525.

\bibitem[Redmon and Farhadi, 2018]{Redmon2018}
Redmon, J. and Farhadi, A. (2018).
\newblock {YOLOv3: An Incremental Improvement}.
\newblock {\em arXiv preprint arXiv:1804.02767}.

\bibitem[Ren et~al., 2015]{Ren2015}
Ren, S., He, K., Girshick, R., and Sun, J. (2015).
\newblock {Faster r-cnn: Towards real-time object detection with region
  proposal networks}.
\newblock In {\em Advances in neural information processing systems}, pages
  91--99.

\bibitem[Rezatofighi et~al., 2019]{Rezatofighi2019}
Rezatofighi, H., Tsoi, N., Gwak, J., Sadeghian, A., Reid, I., and Savarese, S.
  (2019).
\newblock Generalized intersection over union: A metric and a loss for bounding
  box regression.
\newblock In {\em Proceedings of the IEEE Conference on Computer Vision and
  Pattern Recognition}, pages 658--666.

\bibitem[Schroff et~al., 2015]{Schroff2015}
Schroff, F., Kalenichenko, D., and Philbin, J. (2015).
\newblock {FaceNet: A unified embedding for face recognition and clustering}.
\newblock {\em Proceedings of the IEEE Computer Society Conference on Computer
  Vision and Pattern Recognition}, 07-12-June:815--823.

\bibitem[Sohn, 2016]{Sohn2016}
Sohn, K. (2016).
\newblock {Improved deep metric learning with multi-class N-pair loss
  objective}.
\newblock {\em Advances in Neural Information Processing Systems},
  (Nips):1857--1865.

\bibitem[Song et~al., 2020]{Song2020}
Song, G., Liu, Y., and Wang, X. (2020).
\newblock Revisiting the sibling head in object detector.
\newblock In {\em Proceedings of the IEEE/CVF Conference on Computer Vision and
  Pattern Recognition}, pages 11563--11572.

\bibitem[Tian et~al., 2019]{Tian2019}
Tian, Z., Shen, C., Chen, H., and He, T. (2019).
\newblock Fcos: Fully convolutional one-stage object detection.
\newblock In {\em Proceedings of the IEEE international conference on computer
  vision}, pages 9627--9636.

\bibitem[Vicente et~al., 2011]{Vicente2011}
Vicente, S., Rother, C., and Kolmogorov, V. (2011).
\newblock {Object cosegmentation}.
\newblock {\em Proceedings of the IEEE Computer Society Conference on Computer
  Vision and Pattern Recognition}, pages 2217--2224.

\bibitem[Vu et~al., 2019]{Vu2019}
Vu, T., Jang, H., Pham, T.~X., and Yoo, C. (2019).
\newblock Cascade rpn: Delving into high-quality region proposal network with
  adaptive convolution.
\newblock In {\em Advances in Neural Information Processing Systems}, pages
  1432--1442.

\bibitem[Wang et~al., 2018a]{Wang2018b}
Wang, H., Wang, Y., Zhou, Z., Ji, X., Gong, D., Zhou, J., Li, Z., and Liu, W.
  (2018a).
\newblock {CosFace: Large Margin Cosine Loss for Deep Face Recognition}.
\newblock {\em Proceedings of the IEEE Computer Society Conference on Computer
  Vision and Pattern Recognition}, pages 5265--5274.

\bibitem[Wang et~al., 2017]{Wang2017c}
Wang, J., Zhou, F., Wen, S., Liu, X., and Lin, Y. (2017).
\newblock {Deep Metric Learning with Angular Loss}.
\newblock {\em Proceedings of the IEEE International Conference on Computer
  Vision}, 2017-Octob:2612--2620.

\bibitem[Wang et~al., 2018b]{wang2018non}
Wang, X., Girshick, R., Gupta, A., and He, K. (2018b).
\newblock Non-local neural networks.
\newblock In {\em Proceedings of the IEEE conference on computer vision and
  pattern recognition}, pages 7794--7803.

\bibitem[Weber et~al., 2019]{Weber2019}
Weber, M., F{\"{u}}rst, M., and Z{\"{o}}llner, J.~M. (2019).
\newblock {Automated Focal Loss for Image based Object Detection}.
\newblock {\em arXiv preprint arXiv:1904.09048}.

\bibitem[Weinberger and Saul, 2009]{Weinberger2009}
Weinberger, K.~Q. and Saul, L.~K. (2009).
\newblock {Distance metric learning for large margin nearest neighbor
  classification}.
\newblock {\em Journal of Machine Learning Research}, 10:207--244.

\bibitem[Wu and He, 2018]{wu2018group}
Wu, Y. and He, K. (2018).
\newblock {Group normalization}.
\newblock In {\em Proceedings of the European Conference on Computer Vision
  (ECCV)}, pages 3--19.

\bibitem[Xu et~al., 2019]{Xu2019}
Xu, H., Lin, G., and Wang, M. (2019).
\newblock {A Review of Recent Advances in Image Co-Segmentation Techniques}.
\newblock {\em IEEE Access}, 7:182089--182112.

\bibitem[Yang et~al., 2020]{Yang2020}
Yang, Y., Liang, K.~J., and Carin, L. (2020).
\newblock {Object Detection as a Positive-Unlabeled Problem}.
\newblock {\em arXiv preprint arXiv:2002.04672}.

\bibitem[Yuan et~al., 2017]{Yuan2017}
Yuan, Z., Lu, T., and Wu, Y. (2017).
\newblock {Deep-dense conditional random fields for object co-segmentation}.
\newblock {\em IJCAI International Joint Conference on Artificial
  Intelligence}, pages 3371--3377.

\bibitem[Zhang et~al., 2020]{Zhang2019c}
Zhang, S., Chi, C., Yao, Y., Lei, Z., and Li, S.~Z. (2020).
\newblock Bridging the gap between anchor-based and anchor-free detection via
  adaptive training sample selection.
\newblock In {\em Proceedings of the IEEE/CVF Conference on Computer Vision and
  Pattern Recognition}, pages 9759--9768.

\bibitem[Zhang et~al., 2018]{zhang2018single}
Zhang, S., Wen, L., Bian, X., Lei, Z., and Li, S.~Z. (2018).
\newblock {Single-shot refinement neural network for object detection}.
\newblock In {\em Proceedings of the IEEE conference on computer vision and
  pattern recognition}, pages 4203--4212.

\bibitem[Zhang et~al., 2019]{Zhang2019e}
Zhang, Z., He, T., Zhang, H., Zhang, Z., Xie, J., and Li, M. (2019).
\newblock {Bag of Freebies for Training Object Detection Neural Networks}.
\newblock {\em arXiv preprint arXiv:1902.04103}.

\bibitem[Zhou et~al., 2019]{Zhou2019}
Zhou, X., Wang, D., and Kr{\"{a}}henb{\"{u}}hl, P. (2019).
\newblock {Objects as Points}.
\newblock {\em arXiv preprint arXiv:1904.07850}.

\bibitem[Zhu et~al., 2019]{zhu2019deformable}
Zhu, X., Hu, H., Lin, S., and Dai, J. (2019).
\newblock Deformable convnets v2: More deformable, better results.
\newblock In {\em Proceedings of the IEEE Conference on Computer Vision and
  Pattern Recognition}, pages 9308--9316.

\bibitem[Zoph et~al., 2019]{Zoph2019}
Zoph, B., Cubuk, E.~D., Ghiasi, G., Lin, T.-Y., Shlens, J., and Le, Q.~V.
  (2019).
\newblock {Learning Data Augmentation Strategies for Object Detection}.
\newblock {\em arXiv preprint arXiv:1906.11172}.

\end{thebibliography}

\end{document}